\documentclass{article}

\usepackage[T1]{fontenc}

\usepackage{textcomp}
\usepackage[T1]{fontenc}

\usepackage{ijcai19}
\usepackage[utf8]{inputenc}
\usepackage{times}
\usepackage{soul}
\usepackage{url}
\usepackage[hidelinks,breaklinks=true]{hyperref}
\usepackage{xurl}
\usepackage[utf8]{inputenc}
\usepackage[small]{caption}
\usepackage{graphicx}
\usepackage{newunicodechar}
\newunicodechar{₂}{$_2$}
\usepackage{textgreek}
\usepackage{xcolor}
\usepackage{subcaption}
\usepackage{float}
\usepackage{amsmath}
\usepackage{booktabs}
\usepackage{natbib}
\usepackage{amssymb}
\usepackage{csquotes}
\usepackage{multirow} 
\usepackage{longtable}
\usepackage{enumitem}
\usepackage{microtype}
\usepackage{xurl}
\usepackage{doi}
\usepackage{dblfloatfix}
\usepackage{graphicx}
\usepackage[section]{placeins}

\usepackage[font=small,labelfont=bf]{caption}

\title{GPCR-Filter: a deep learning framework for efficient and precise GPCR modulator discovery}

\author{
Jingjie Ning$^{2}$\thanks{These authors contributed equally to this work.}\and
Xiangzhen Shen$^{3}$\footnotemark[1]\and
Li Hou$^{1}$\footnotemark[1]\and
Shiyi Shen$^{1}$\and
Jiahao Yang$^{4}$\and
Junrui Li$^{1}$\and
Hong Shan$^{1}$\and
Sanan Wu$^{5}$\and
Sihan Gao$^{6}$\and
H. Eric Xu$^{1}$\thanks{Corresponding authors. Email: xinheng.he@simm.ac.cn, eric.xu@simm.ac.cn.}\and
Xinheng He$^{1}$\footnotemark[2]
\affiliations
$^{1}$ The State Key Laboratory of Drug Research, Shanghai Institute of Materia Medica,
Chinese Academy of Sciences, Shanghai, China\\
$^{2}$ School of Computer Science, Carnegie Mellon University, Pittsburgh, PA 15213, USA\\
$^{3}$ College of Computer Science and Electronic Engineering, Hunan University, Changsha, Hunan 410082, China\\
$^{4}$ Lingang Laboratory, Shanghai, China\\
$^{5}$ Research Center for Medicinal Structural Biology, National Research Center for Translational Medicine at Shanghai,
State Key Laboratory of Medical Genomics, Ruijin Hospital, Shanghai Jiao Tong University School of Medicine, Shanghai, China\\
$^{6}$ School of Pharmacy, Fudan University, Shanghai 201203, China
}

\begin{document}

\maketitle
\pagenumbering{arabic}

\begin{abstract}
G protein–coupled receptors (GPCRs) govern diverse physiological processes and are central to modern pharmacology. Yet discovering GPCR modulators remains challenging because receptor activation often arises from complex allosteric effects rather than direct binding affinity, and conventional assays are slow, costly, and not optimized for capturing these dynamics. Here we present GPCR-Filter, a deep learning framework specifically developed for GPCR modulator discovery. We assembled a high-quality dataset of over 90,000 experimentally validated GPCR–ligand pairs, providing a robust foundation for training and evaluation. GPCR-Filter integrates the ESM-3 protein language model for high-fidelity GPCR sequence representations with graph neural networks that encode ligand structures, coupled through an attention-based fusion mechanism that learns receptor–ligand functional relationships. Across multiple evaluation settings, GPCR-Filter consistently outperforms state-of-the-art compound–protein interaction models and exhibits strong generalization to unseen receptors and ligands. Notably, the model successfully identified micromolar-level agonists of the 5-HT\textsubscript{1A} receptor with distinct chemical frameworks. These results establish GPCR-Filter as a scalable and effective computational approach for GPCR modulator discovery, advancing AI-assisted drug development for complex signaling systems.
\end{abstract}

\section{Introduction}
G protein–coupled receptors (GPCRs) constitute the largest and most diverse family of cell-surface receptors, with more than 800 members that orchestrate a wide spectrum of physiological and pathological processes\cite{munk2019online}. Through coupling with G proteins and arrestins, GPCRs regulate key pathways in the nervous, endocrine, and immune systems\cite{mueller2024empirical, duan2024cryo, dickinson2024gpcrs}. Owing to their central roles in signaling transduction, GPCRs constitute one of the most important classes of drug targets in modern medicine: up to 36\% of all drugs approved by the U.S. Food and Drug Administration (FDA) act on GPCRs, and more than 300 GPCR-targeting agents are currently in clinical development\cite{hauser2017trends, lorente2025gpcr}. The discovery of new GPCR modulators is increasingly driven by the integration of computational and experimental approaches\cite{michino2025ai}.

Despite decades of research, identifying GPCR modulators remains a major challenge\cite{zhang2024g}. Unlike enzymes or ion channels, GPCRs exhibit highly dynamic conformational landscapes and allosteric mechanisms that can decouple ligand binding from downstream signaling efficacy\cite{peter2024comparative, lu2021activation}. Consequently, traditional computational screening methods often fail to capture the complex relationship between binding and functional modulation, resulting in uncertain or inactive hits, especially in unpublished negative results\cite{ballante2021structure}. Experimental assays designed to probe these effects are typically costly, low-throughput, and labor-intensive, further limiting the discovery rate of novel modulators\cite{guo2022recent}.

Recent advances in artificial intelligence (AI) offer new opportunities to address these challenges. AI models are capable of learning the complex features and approaching ligand discovery in a data-driven manner. Broadly, AI-based methods can be categorized into structure-based and sequence-based approaches. Structure-based strategies leverage three-dimensional receptor conformations to extract pocket information, filter potential binders, and capture detailed molecular interactions through AI-assisted docking\cite{lu2024drugclip, sim2025recent}. However, the number of experimentally determined GPCR–ligand complex structures remains limited (fewer than 1,800, and largely conserved), and the inaccuracies of computationally predicted structures constrain the ability to fine-tune such models effectively\cite{GPCRdb, shen2025update}.

In contrast, sequence-based screening approaches offer access to far larger datasets and enable modeling of the relationships between ligand chemistry, receptor sequence, and downstream functional activity. General sequence-based drug–target interaction (DTI) models such as TransformerCPI\cite{chen2020transformercpi}, TransformerCPI2.0\cite{chen2023sequence}, and ConPLex\cite{singh2023contrastive} combine protein language models with molecular representations to predict binding likelihoods across vast chemical spaces. Nonetheless, these frameworks are primarily optimized for generic binding prediction and often lack the biochemical sensitivity required to capture the nuanced sequence–structure–function relationships that govern GPCR pharmacology. Combining sequence and structure based screening method, especially in GPCR domain, can improve success rate via different aspects \cite{ballante2021structure}.

To overcome these limitations and provide a new filter during screening, we curated a large, experimentally grounded dataset comprising over 90,000 GPCR–ligand pairs. Based on this dataset, we developed GPCR-Filter, a deep learning framework dedicated to GPCR modulator discovery. GPCR-Filter integrates the ESM-3 pre-trained model\cite{esm_3} for fine-grained GPCR sequence embeddings with graph neural networks (GNNs) that capture ligand structure and chemistry. An attention-based fusion mechanism learns functional correspondence between receptor and ligand features, enabling prediction of modulatory potential beyond simple binding.

Across multiple evaluation scenarios, GPCR-Filter surpasses state-of-the-art DTI models in both accuracy and generalization. Furthermore, experimental validation demonstrated that GPCR-Filter can identify micromolar-level agonists of the 5-HT\textsubscript{1A} receptor with distinct scaffolds, confirming its potential to guide functional ligand discovery. By bridging GPCR biology and deep learning, this work provides a scalable computational framework for mapping modulator space and accelerating GPCR-targeted drug development.

\section{Results}

\subsection{GPCR-Filter overview}
As illustrated in Fig. 1a, GPCR modulator discovery typically begins with selecting a GPCR target and assembling a ligand library suitable for virtual screening. Following conventional structure-based filtering, GPCR-Filter operates purely on sequence and SMILES information to further prioritize candidate ligands before experimental testing, thereby increasing downstream hit rates. The overall architecture is shown in Fig. 1b. It takes a GPCR amino-acid sequence and a ligand SMILES string as input, encoding them into per-residue and per-atom representations using the pretrained protein language model ESM-3 and a molecular graph neural network, respectively. These representations are projected into a shared latent space and fused through a Transformer-style decoder with ligand-to-protein cross-attention, which aggregates the coupled features into a final interaction probability. 

To train and evaluate GPCR-Filter, we curated a large, high-quality dataset of 91,396 validated human GPCR–drug interactions, integrating records from GPCRdb and GtoPdb\cite{GPCRdb, GtoPdb}, aligning 527 unique GPCRs to UniProt sequences, and standardizing 72,177 distinct ligands into canonical SMILES. The interaction distribution exhibits a pronounced long-tailed pattern in which a small number of highly promiscuous GPCRs dominate the dataset, whereas most receptors exhibit sparse ligand coverage (Appendix Fig. S1 and Table S1). Because negative labels are not experimentally available, we constructed negatives by enumerating all GPCR–ligand combinations, removing known positives, and sampling to obtain a balanced 1:1 dataset. Across these settings, we compared GPCR-Filter with two state-of-the-art sequence-based DTI models, ConPLex and TransformerCPI2.0, using identical sequence+SMILES inputs; detailed dataset statistics and split sizes are provided in Methods\cite{chen2023sequence, singh2023contrastive}.

\begin{figure*}[tbp]
  \centering
  \includegraphics[width=1.0\textwidth]{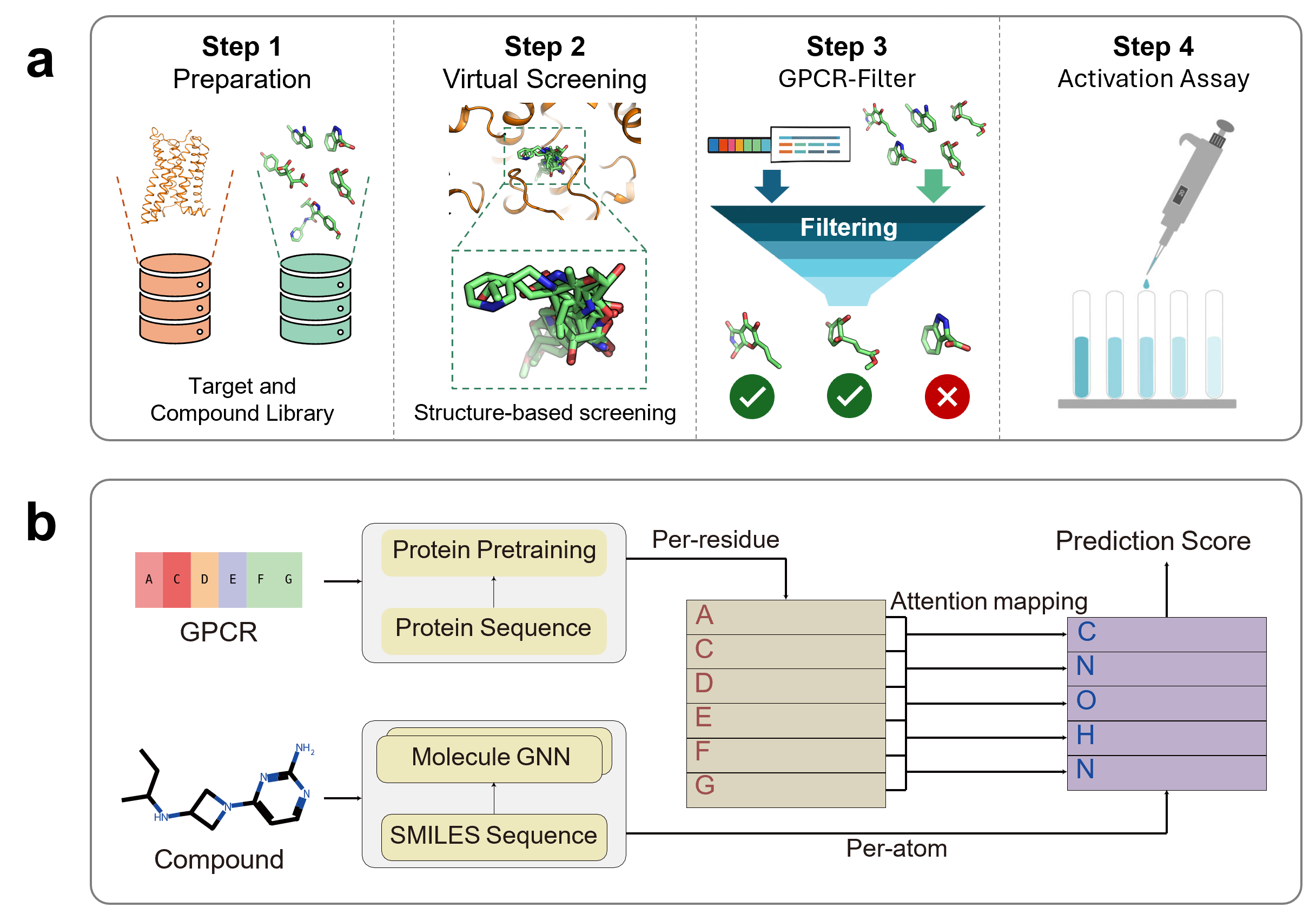}
  \caption{\textbf{Application of GPCR-Filter and overview of its architecture.}
  \textbf{(a)} Workflow illustrating how GPCR-Filter integrates into GPCR modulator discovery. After ligand preparation and initial structure-based virtual screening, GPCR-Filter uses only GPCR sequences to further refine docking outputs and prioritize candidates for downstream activation assays. 
  \textbf{(b)} Schematic of the GPCR-Filter architecture. GPCR sequences are embedded into per-residue representations using a pretrained protein language model, while ligand SMILES are encoded into per-atom features through a molecular graph neural network. A cross-attention module couples ligand and receptor representations, and the fused features are aggregated to produce a final interaction probability.}
  \label{fig:architecture}
\end{figure*}

\subsection{Predictive performance across data splits}

We compared GPCR-Filter with two competitive sequence-based DTI frameworks, ConPLex and TransformerCPI2.0, under three evaluation regimes designed to probe increasingly challenging levels of generalization (Fig. 2a). These include a random split for in-distribution assessment, an intra-target split evaluating generalization to unseen ligand combinations for previously encountered GPCRs, and an inter-target split testing cross-target transfer to entirely unseen receptors (see Methods for details). We used accuracy (ACC), area under the ROC curve (AUC), average precision (AP) and precision as evaluation metrics, where larger values indicate better performance. As shown in Table \ref{tab:combined_performance}, GPCR-Filter achieved near-ceiling performance under the random split (AUC of 98.93\% and AP of 98.70\%), substantially outperforming the two baselines. Its robustness was further reflected in the intra-target setting (Table \ref{tab:combined_performance}), where GPCR-Filter maintained high discriminative power (AUC of 97.16\% and AP of 96.86\%) despite the increased difficulty of predicting unseen ligand combinations for the same receptor. By contrast, both ConPLex and TransformerCPI2.0 showed consistently low AUCs (all below 63\%), suggesting limited capacity to generalize across ligand space.

We next examined the inter-target scenario, the most stringent generalization task in which receptors in the test set never appear during training. In this setting (Table \ref{tab:combined_performance}), GPCR-Filter again delivered the strongest performance, with an AUC of 73.44\% and AP of 64.04\%, markedly higher than TransformerCPI2.0 and substantially above ConPLex, whose AUC fell below 50\%. Such an AUC value shows ConPLex may have overfitted on its training set. This indicates that GPCR-Filter is more capable of capturing receptor-level sequence determinants relevant for ligand recognition. Collectively, these results demonstrate that GPCR-Filter consistently outperforms established DTI baselines across all three evaluation regimes, with particularly pronounced advantages in out-of-distribution settings requiring transfer across ligand chemical space or GPCR sequence space.

\begin{figure*}[tbp]
  \centering
  \includegraphics[width=0.8\textwidth]{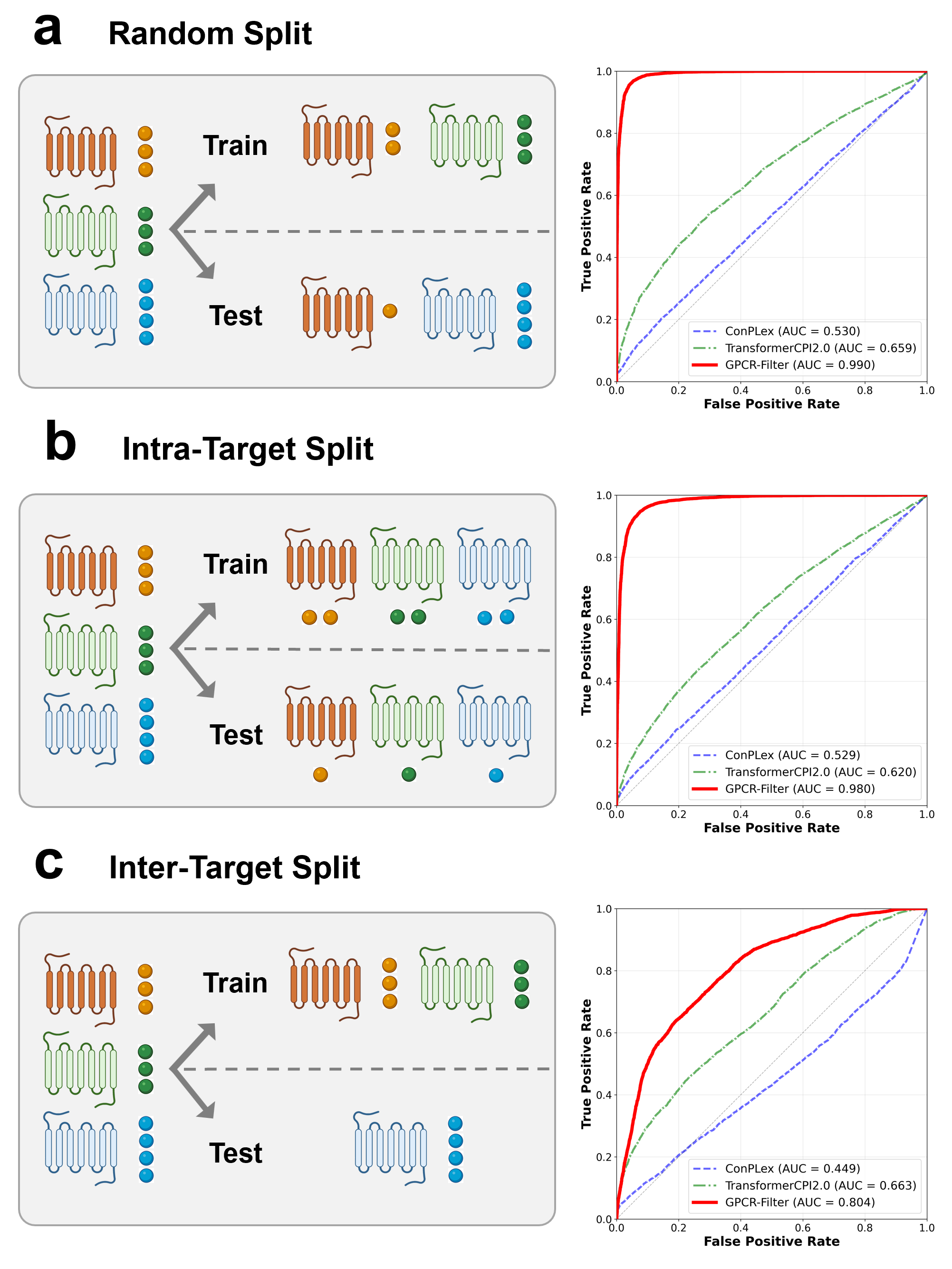}
  \caption{\textbf{Dataset split strategies and performance overview.}
  \textbf{(a)} Random Split: Each GPCR (receptor icon) and its ligands are randomly assigned to training and testing sets. Both receptors and ligands can appear in both sets, simulating a fully mixed in-distribution evaluation.  
  \textbf{(b)} Intra-Target Split: Each receptor is present in both training and testing, but its associated ligands are divided into mutually exclusive subsets. This evaluates generalization to unseen ligands for the same receptor.  
  \textbf{(c)} Inter-Target Split: The receptor set is partitioned into disjoint training and held-out target subsets (e.g., 9:1). Validation examples are drawn from the held-out subset to keep target identities disjoint between training and evaluation.  
  On the right, the ROC curves illustrate the discriminative performance of GPCR-Filter and baseline models under the three data partitioning protocols. Receptor and ligand icons are for schematic illustration only, with color denoting training or testing samples.}
  \label{fig:split_schematic_with_roc}
\end{figure*}

We further visualized the three partitioning strategies and model performance using ROC curves (Fig. \ref{fig:split_schematic_with_roc}), which illustrate the strong separation achieved by GPCR-Filter across all protocols. Together, these findings highlight the robustness and generalizability of GPCR-Filter for GPCR–ligand interaction prediction under diverse distributional shifts.

\begin{figure*}[tbp]
  \centering
  \includegraphics[width=0.9\textwidth]{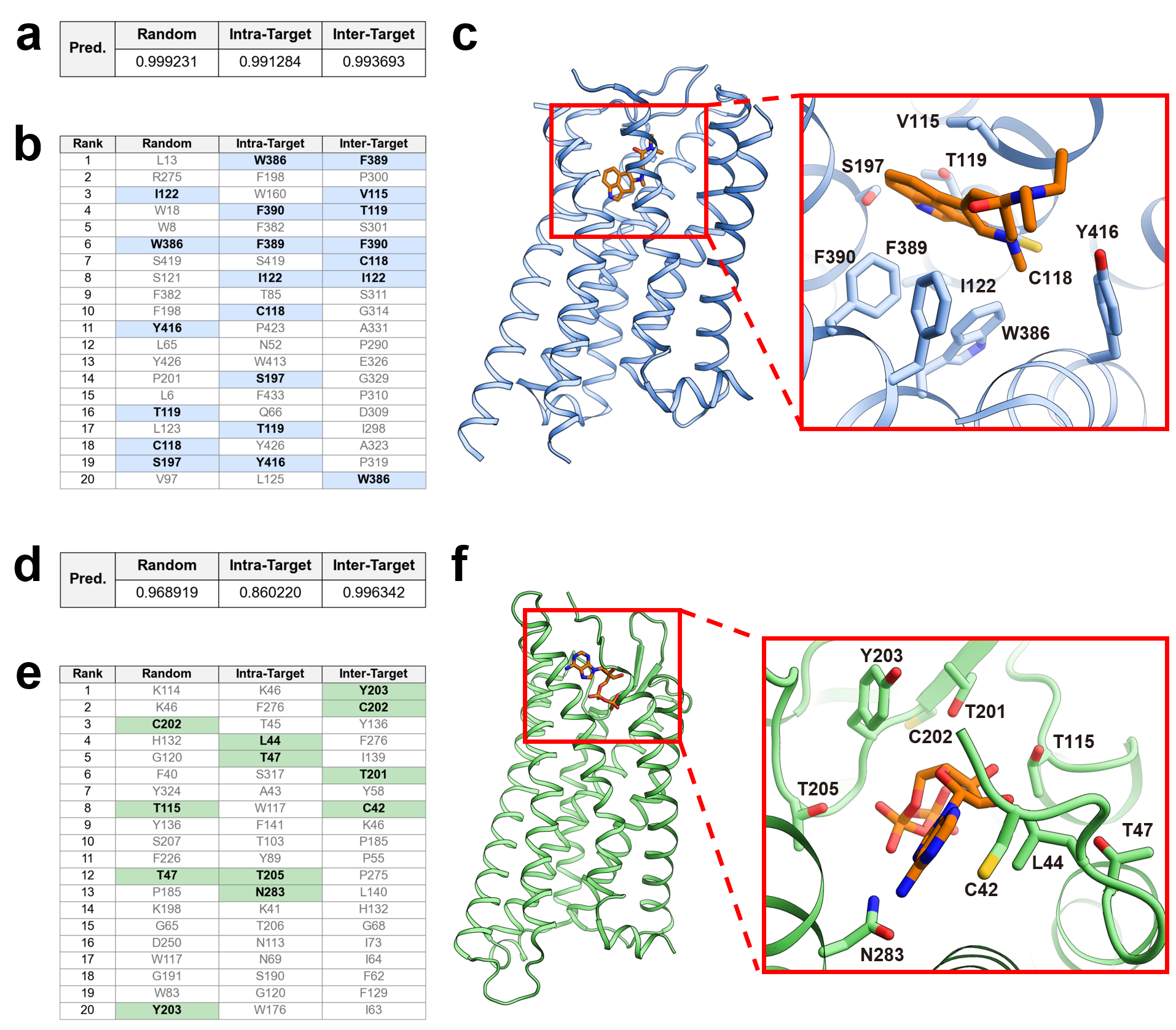}
  \caption{\textbf{Binding-pocket analysis across two GPCR complexes.}
  \textbf{(a–c)} correspond to \textbf{PDB 9bsb}; \textbf{(d–f)} correspond to \textbf{PDB 9jcl}.
    \textbf{(a,d)} Model prediction scores under the three evaluation settings (Random, Intra-target, Inter-target). All predicted probabilities exceed 0.5, indicating consistent positive predictions across splits. \textbf{(b,e)} Top-20 attended residues ranked under the three evaluation settings. Each entry is reported as \emph{index–residue}; highlighted cells indicate residues that are repeatedly prioritized across splits. \textbf{(c,f)} Spatial mapping of the \textit{Inter-Target} model’s Top-20 attended residues onto the corresponding structure and binding pocket. 
    Inset panels show the ligand in sticks and the pocket region enlarged (red box), illustrating which high-attention residues lie within the crystallographic binding site. \emph{All residue indices correspond to the numbering used in the PDB sequences.}
}

  \label{fig:pocket_analysis}
\end{figure*}

\begin{table*}[t]
\centering
\renewcommand{\arraystretch}{1.4}
\caption{\textbf{Performance comparison of GPCR-Filter and baseline models across the three evaluation regimes.}}
\begin{tabular}{lcccccccccccc}
\toprule
\multirow{2}{*}{\textbf{Method}} 
& \multicolumn{4}{c}{\textbf{Random}} 
& \multicolumn{4}{c}{\textbf{Intra-target}} 
& \multicolumn{4}{c}{\textbf{Inter-target}} \\
\cmidrule(lr){2-5} \cmidrule(lr){6-9} \cmidrule(lr){10-13}
& ACC & AUC & AP & Prec 
& ACC & AUC & AP & Prec
& ACC & AUC & AP & Prec \\
\midrule
ConPLex 
& 52.82 & 53.03 & 55.29 & 57.66
& 52.01 & 52.89 & 55.26 & 56.92
& 57.85 & 44.91 & 40.78 & 39.93 \\

TransformerCPI2.0 
& 61.46 & 65.94 & 68.10 & 62.50
& 58.14 & 62.02 & 63.34 & 58.77
& 63.33 & 66.31 & 58.50 & 53.35 \\

\textbf{GPCR-Filter} 
& \textbf{95.68} & \textbf{99.02} & \textbf{98.87} & \textbf{95.19}
& \textbf{93.53} & \textbf{97.98} & \textbf{97.78} & \textbf{92.72}
& \textbf{74.57} & \textbf{80.35} & \textbf{70.99} & \textbf{75.93} \\
\bottomrule
\end{tabular}
\label{tab:combined_performance}
\end{table*}


\begin{figure*}[tb]
  \centering
  \includegraphics[width=1.0\textwidth]{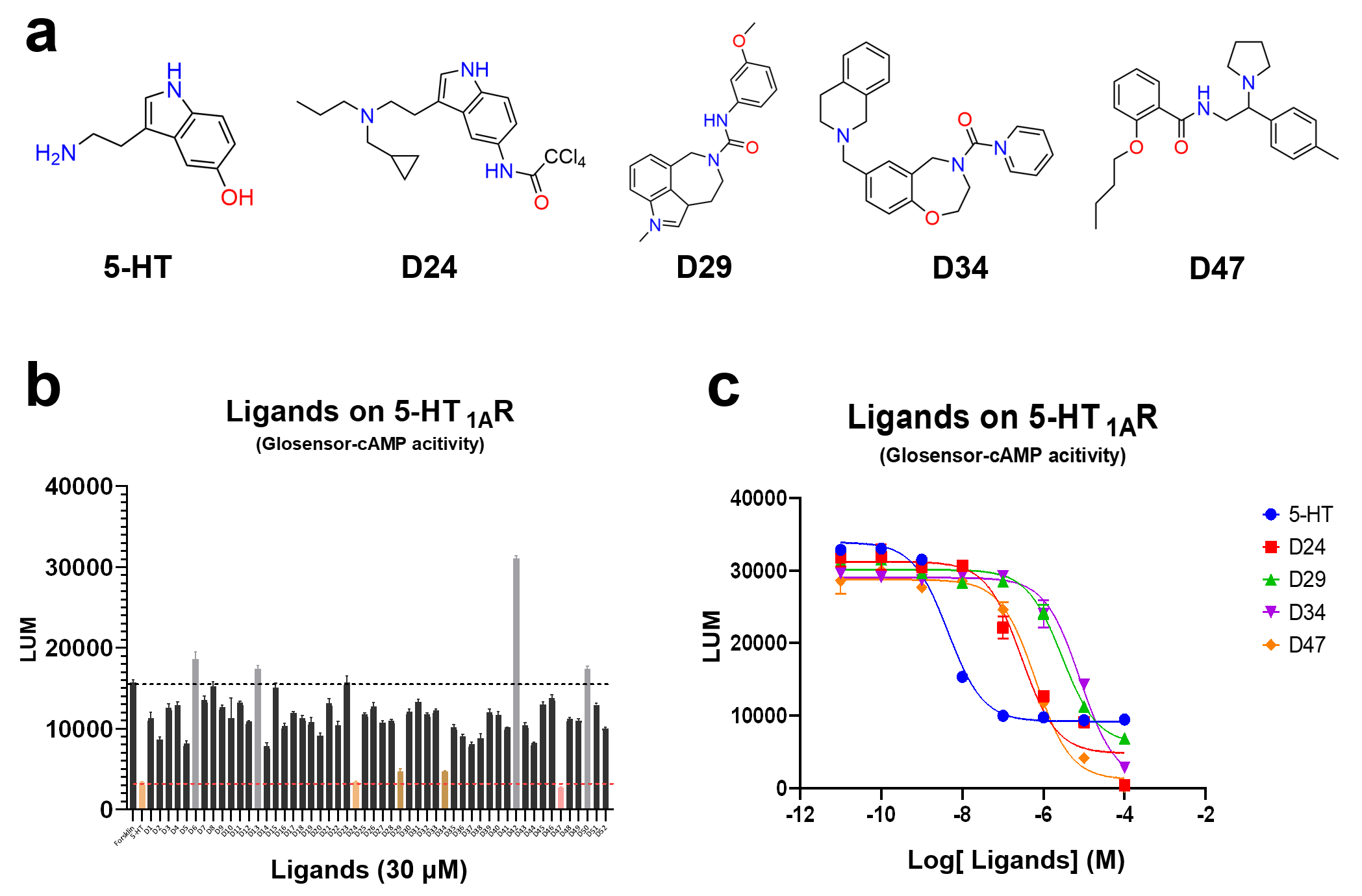}
  \caption{\textbf{Experimental validation of GPCR-Filter-predicted agonists on the 5-HT\(_{1\mathrm{A}}\) receptor.}
  \textbf{(a)} Chemical structures of 5-HT and the four validated hits.
  \textbf{(b)} Single-concentration screening at 30~\(\mu\)M using a GloSensor-cAMP assay. Bars represent luminescence (LUM) values for individual compounds; dashed lines denote Forskolin (black, baseline) and 5-HT (red, canonical agonist). Four compounds prioritized by GPCR-Filter (D24, D29, D34, D47) exhibited reduced LUM, indicative of receptor activation.
  \textbf{(c)} Normalized concentration--response curves for 5-HT and the four hits, showing comparable or higher maximal effects (E\(_\mathrm{max}\)) but right-shifted potencies (higher EC\(_{50}\)), indicating increased efficacy but lower potency relative to 5-HT.
  Together, these results confirm that GPCR-Filter successfully identified four true agonists activating 5-HT\(_{1\mathrm{A}}\) in a dose-dependent manner.}
  \label{fig:5ht_validation}
\end{figure*}

\subsection{Interpretability of GPCR-Filter}

To understand why GPCR-Filter generalizes and what patterns it relies on, we performed two complementary interpretability analyses. At the dataset level, we characterized each GPCR by the chemical distribution of its known binders. For the top 20 GPCRs with sufficient ligand data, each ligand was encoded using a 2048-bit Morgan (ECFP4) fingerprint \cite{fingerprint}, and a receptor-level representation was obtained by averaging the fingerprints of all ligands associated with that receptor. Pairwise receptor similarity was estimated using Tanimoto distance \cite{Bajusz_Rácz_Héberger_2015} between these averaged profiles, and hierarchical clustering (average linkage) produced a dendrogram and a ligand-profile similarity heatmap (Fig.~\ref{fig:clustering}). Receptors with chemically similar modulators group together in this space.

Although ligand-profile similarity is an indirect proxy for receptor structural similarity, this analysis highlights that GPCRs occupy partially organized regions of chemical space. Such organization provides a plausible context for cross-target generalization: if GPCR-Filter learns chemically grounded patterns shared across subsets of receptors, then unseen receptors whose ligand chemistry resembles that of training receptors may fall within regions where these learned patterns remain applicable. Thus, the clustering results offer an interpretive framework for understanding how transferable chemical preferences could support inter-target predictive performance.

At the structural level, we focused on whether the model has learned transferable binding principles rather than merely memorizing receptor–ligand pairs. Because cross-attention plays a central role in ligand–protein interaction modeling, we assess whether residues receiving high attention weights correspond to crystallographic pocket residues \cite{nazem2024deep, zhang2025labind}. Specifically, we extract the decoder’s last-layer ligand-to-protein cross-attention using the ligand CLS token ($t=0$) as query and residue embeddings as keys/values, average across heads, mask padded positions, L1-normalize the resulting residue-wise vector, and rank residues to obtain a Top-20 list. Crystallographic pocket residues are defined as those whose heavy-atom distance to the ligand is $<5$\,\AA{}. For each PDB structure, we report the number of pocket residues appearing in the Top-20 (Pocket hits@20). Models trained under random, intra-target, and inter-target protocols yield broadly consistent top hits, suggesting that the attention patterns are stable across training regimes and provide a useful signal for pocket localization.

We illustrate these patterns using two recently solved GPCR–ligand complexes: the dopamine D\(_2\) receptor (DRD2; PDB \texttt{9bsb} \cite{9bsb}) and the purinergic receptor (PDB \texttt{9jcl} \cite{9jcl}). In both cases, GPCR-Filter confidently predicts the co-crystallized ligand as active under all three training protocols (Fig.~\ref{fig:pocket_analysis}a,d), providing suitable examples for qualitative interpretability. For \texttt{9bsb} (Fig.~\ref{fig:pocket_analysis}a–c), there are nine pocket residues within 5 Å of the ligand. The Top-20 attention ranks recover \textbf{6}, \textbf{8}, and \textbf{7} pocket residues for the random, intra-target, and inter-target models, respectively, indicating that attention emphasizes residues mediating direct interactions. For \texttt{9jcl} (Fig.~\ref{fig:pocket_analysis}d–f), we observe a similar qualitative trend: pocket residues are consistently enriched among the Top-20 attention positions across all splits, again showing that attention tends to focus on residues involved in diverse ligand–receptor interactions. Taken together, these case studies demonstrate that the cross-attention mechanism preferentially highlights experimentally validated binding-site residues, supporting that GPCR-Filter captures protein–ligand interaction patterns through pocket-relevant attention signals rather than memorizing receptor–ligand pairs.

\subsection{Wet-lab validation on 5-HT\texorpdfstring{$_{1\mathrm{A}}$}{1A} receptor}

To validate GPCR-Filter in wet-lab experiments and assess its ability to filter GPCR modulators, a virtual screening pipeline was established. In this pipeline, 1{,}644{,}833 ChemDiv compounds were docked into the SEP-363856 binding pocket of the 5-HT$_{1\mathrm{A}}$ receptor (PDB~ID:~8W8B). The top-ranked docking candidates (8{,}705 molecules) were then screened using GPCR-Filter; among them, 97 compounds with predicted binding probability \(p>0.5\) were selected for purchase, and, due to availability, 52 were obtained and tested. We conducted in-vitro validation of these 52 GPCR-Filter-prioritized compounds targeting the human 5-HT$_{1\mathrm{A}}$ receptor using a GloSensor-cAMP assay. In this readout, lower luminescence (LUM) indicates stronger receptor activation. Forskolin served as a high-LUM reference, and 5-HT as an agonist control.

We first performed a single-concentration screen at 30~\(\mu\)M for the 52 compounds prioritized by GPCR-Filter. Four ligands (D24, D29, D34, D47) produced more pronounced LUM reductions compared to that of 5-HT, indicating strong receptor activation. These four ligands were then selected for multiple-concentration pharmacological characterization.

For D24, D29, D34, and D47, concentration--response curves were measured and normalized to the 5-HT maximal effect (100\%). All four compounds elicited sigmoidal inhibition curves consistent with Gi-coupled receptor activation. Their maximal effects ($E_{\mathrm{max}}$) approached or exceeded that of 5-HT, though with lower apparent potency as reflected by right-shifted $\mathrm{EC}_{50}$ values around the micromolar level, suggesting that these compounds represent promising hits for affinity optimization.

\section{Discussion}

In this work, we presented GPCR-Filter, a GPCR-focused interaction model trained on a curated human GPCR–ligand dataset. On this benchmark, the model delivers strong performance under random, intra-target, and inter-target splits and maintains a clear margin under the most stringent inter-target setting (Tab.~\ref{tab:combined_performance}), consistent with the designed difficulty gradient of these protocols (Fig.~\ref{fig:split_schematic_with_roc}). Wet-lab assays on 5-HT$_{1\mathrm{A}}$ validated four agonists among the top-ranked predictions (Fig.~\ref{fig:5ht_validation}), and attention-based analyses align model decisions with crystallographic binding regions (Fig.~\ref{fig:pocket_analysis}). Because near-ceiling performance under the random split (AUC $\approx 0.99$) can partly reflect the memorization-prone nature of random partitioning and potential leakage, we further evaluated a Bemis–Murcko scaffold split and a GPCR-family hold-out protocol to more rigorously assess generalization. Together, these results support GPCR-Filter as a practical in-silico screening component for GPCR discovery pipelines.

Across all three evaluation regimes, GPCR-Filter exhibits strong representational and predictive capability. Notably, in the inter-target regime ConPLex exhibits AUC below 0.5 while maintaining accuracy above 50\%, likely reflecting overfitting inside their model evaluation. We attribute this robustness to a combination of high-quality data and multi-level model design. The curated GPCR–ligand dataset spans diverse targets and chemotypes, uses balanced supervision, and applies carefully controlled split strategies. At the dataset level, GPCRs display structured organization in ligand-profile space (Fig.~\ref{fig:clustering}), providing a plausible basis for cross-target transfer by placing unseen receptors near regions where learned chemical patterns remain applicable. On the modeling side, GPCR-Filter integrates a pretrained protein language model (ESM-3) with a lightweight graph neural network over ligand atoms, coupled through coordinated self- and ligand→protein cross-attention. These mechanisms allow the decoder to focus on functionally relevant residues: Top-20 attention-ranked residues are enriched near crystallographic pockets across two newly solved complexes and across all split types (Fig.~\ref{fig:pocket_analysis}), yielding residue-level maps that offer mechanistic interpretability. This synergy between chemically structured data and attention-based fusion supports high predictive accuracy even under challenging out-of-distribution settings.

Despite these strengths, several limitations remain. Negative labels are obtained through 1:1 negative sampling and inevitably include unobserved actives; exploring alternative decoy strategies, sampling ratios, and re-sampling variance may further clarify performance bounds. In the inter-target split, ligands are allowed to recur across different targets; although the receptors themselves are disjoint, ligand reuse may provide an easier signal for certain chemotypes, and stricter ligand or scaffold-level de-duplication could further stress-test the model. Also, broader comparisons involving deeper calibration strategies or structure-aware models remain to be explored.

Overall, this study demonstrates that combining curated, GPCR-specific data with multi-level fusion of protein language models and ligand graph networks yields a data-efficient and interpretable virtual screening model with strong out-of-distribution performance. Future work will expand validation to additional protein families, refine negative sampling and calibration protocols, and incorporate structural or pocket-level constraints to further strengthen inter-target generalization. More broadly, the results suggest that chemically grounded data organization—combined with attention mechanisms that expose ligand–receptor interaction patterns—offers a promising foundation for next-generation GPCR drug discovery pipelines.

\section{Methods}

\subsection{Overview of GPCR-Filter}
GPCR-Filter predicts GPCR--ligand interactions from a GPCR amino acid sequence and a ligand SMILES string. It first produces per-residue and per-atom embeddings, projects them to a shared hidden size, and then fuses them via a Transformer-style cross-attention decoder to yield a binary interaction score (see \autoref{fig:architecture} for a schematic). Concretely, the GPCR side is encoded by ESM-3 encoder for precomputed per-residue embeddings, while the ligand side is embedded by a linear layer plus a single GCN convolution \cite{semisupervisedclassificationgraphconvolutional} and then fed as the \emph{decoder} target sequence; the decoder attends from ligand tokens to encoded GPCR tokens. The graph-level prediction is read from the ligand sequence’s first (graph-level) token.

\subsection{Problem definition}
We formulate the task on tuples \((D, T)\), where \(D\) is a ligand and \(T\) is a GPCR. The model outputs 2-way logits \(\mathbf{o}(D,T)\in\mathbb{R}^2\); the positive-class probability is \(p=\mathrm{softmax}(\mathbf{o})_{[1]}\). During training we minimize the binary cross-entropy implemented as 2-class cross-entropy over logits; at evaluation time, unless otherwise specified, threshold-dependent metrics (ACC/Precision/Recall/F1) use a fixed probability threshold of \(0.5\), while AUC/AUPR are threshold-free.

\subsection{GPCR embedding calculation}
For GPCRs, we start from precomputed per-residue representations (e.g., ESM-3 outputs \cite{esm_3}), of dimension \(h_t\) (in our runs \(h_t{=}1536\)). A linear projection \(\mathbb{R}^{h_t}\!\to\!\mathbb{R}^{d}\) maps each residue to the shared hidden size \(d\) (we use \(d{=}256\)). The sequence is then encoded by a Transformer \textbf{encoder} with \(L\) layers (default \(L{=}2\)), each layer using 8 attention heads, a feed-forward width of \(4d\), dropout \(0.1\), and standard residual/LayerNorm; a padding mask is applied to ignore padded residues. This encoded sequence serves as the \emph{memory} for cross-attention in the decoder.

\subsection{Ligand embedding calculation}
Ligand SMILES are converted into molecular graphs \(G=(V,E)\) via RDKit with standard atom/bond descriptors.
Node features of size \(h_d\) are first linearly projected to the shared hidden size \(d\), then passed through a single Graph Convolutional Network (GCN) layer.
At layer \(\ell\),

\begin{equation}
\mathbf{x}^{(\ell+1)}_{d,i}
\;=\;
\sigma\!\left(
\sum_{v_j \in \mathcal{N}(v_i)\cup\{v_i\}}
\frac{1}{c_{ij}}\,
\mathbf{W}^{(\ell)} \mathbf{x}^{(\ell)}_{d,j}
\right),
\end{equation}

where \(\mathbf{x}^{(\ell)}_{d,i}\) denotes the node feature of atom \(i\) at layer \(\ell\), \(\mathbf{W}^{(\ell)}\) is the trainable weight matrix, \(\sigma(\cdot)\) is ReLU, and \(c_{ij}\) is a degree-based normalization term.
After GCN, we \emph{prepend a learnable graph-level token} (used for graph-level readout) to the atom sequence and pad to the batch maximum length to form the target sequence
\(\tilde{\mathbf{X}}_{d} \in \mathbb{R}^{(1+|V|)\times d}\) with a corresponding target-side padding mask (see Decoder for fusion).

\subsection{Decoder and attention-based fusion}
At this stage, we have per-residue GPCR features $x_t \in \mathbb{R}^{|S| \times d}$ (after linear projection and an \emph{encoder}) and per-atom ligand features $x_d \in \mathbb{R}^{|V| \times d}$. 
Before fusion, we \emph{prepend a learnable graph-level token} to the ligand sequence and pad to batch length; with this convention we overload notation and still write $x_d \in \mathbb{R}^{(1{+}|V|) \times d}$, where index $0$ is the graph-level (CLS-like) token. 
To perform feature fusion and generate the final prediction, we adopt a Transformer-based \emph{decoder} architecture \cite{NIPS2017_attention}: the ligand sequence $x_d$ serves as the target/query, while the encoded protein sequence $x_t$ serves as the memory (key/value). 
Multi-head scaled dot-product attention with residual connections, LayerNorm, dropout, and padding masks is used in all sublayers (equations below are single-head forms for compactness).

First, the ligand target undergoes self-attention to refine its representation, and the protein sequence is refined by encoder self-attention (written here in the same notation for brevity).
In the self-attention mechanism, each feature vector attends to others within the same set:
\begin{equation}
\alpha_{ij} = \frac{\exp{(x_{d_i} W_q \cdot x_{d_j} W_k^T)}}{\sum_{j'} \exp{(x_{d_i} W_q \cdot x_{d_{j'}} W_k^T)}},
\end{equation}
where $\alpha_{ij}$ denotes the attention weight between ligand features $x_{d_i}$ and $x_{d_j}$, and $W_q, W_k$ are learnable projections. The updated ligand features are
\begin{equation}
x_{d_i}' = \sum_j \alpha_{ij} (x_{d_j} W_v) + x_{d_i}, \quad x_d' \in \mathbb{R}^{(1{+}|V|) \times d}.
\end{equation}
Similarly, protein features are updated by encoder self-attention:
\begin{equation}
\beta_{ij} = \frac{\exp{(x_{t_i} W_q \cdot x_{t_j} W_k^T)}}{\sum_{j'} \exp{(x_{t_i} W_q \cdot x_{t_{j'}} W_k^T)}},
\end{equation}
\begin{equation}
x_{t_i}' = \sum_j \beta_{ij} (x_{t_j} W_v) + x_{t_i}, \quad x_t' \in \mathbb{R}^{|S| \times d},
\end{equation}
where $W_v$ is another trainable projection.

Next, the ligand acts as the query and the protein as key/value in cross-attention. The attention scores are
\begin{equation}
\gamma_{ij} = \frac{\exp{(x_{d_i}' W_q \cdot x_{t_j}' W_k^T)}}{\sum_{j'} \exp{(x_{d_i}' W_q \cdot x_{t_{j'}}' W_k^T)}},
\end{equation}
and the ligand features are updated by aggregating protein information:
\begin{equation}
x_{d_i}'' = \sum_j \gamma_{ij} (x_{t_j}' W_v) + x_{d_i}', \quad x''_d \in \mathbb{R}^{(1{+}|V|) \times d}.
\end{equation}

Finally, we extract the graph-level token at index $0$, $x''_d[0] \in \mathbb{R}^{d}$, and pass it through a multi-layer perceptron (MLP) to obtain the final prediction (two-class logits; we denote the scalar score by $s$ for brevity). 
This attention pathway enables the model to capture ligand–GPCR interactions by dynamically integrating information from relevant regions on both sides; the MLP maps the fused representation to the interaction likelihood. 
For interpretability, we use the \emph{last-layer} cross-attention weights from the ligand graph-level token to residues as residue-level importance scores.

\subsection{Dataset curation}
\label{sec:dataset_curation}
We curated a large-scale, high-quality dataset of human drug--GPCR interactions to enable reliable supervised learning. Human GPCR- and ligand-related records were collected from \textbf{GPCRdb}~\cite{GPCRdb} and \textbf{GtoPdb}~\cite{GtoPdb}. After deduplication, we enumerated the sets of GPCRs and drugs to obtain in total \textbf{91,396} validated drug--GPCR interaction records. GPCR identifiers were aligned to \textbf{UniProt}~\cite{UniProt}, yielding \textbf{527} distinct GPCR FASTA sequences; drug identifiers were re-encoded to obtain \textbf{72,177} distinct drug SMILES strings.

\subsection{Dataset distribution analysis}
\label{sec:dataset_distribution}
Because the ligand set is much larger than the GPCR set, individual receptors are associated with very different numbers of ligands. We quantified this imbalance by examining how frequently each GPCR appears in the interaction dataset. The overall distribution is presented as a \emph{binned} frequency-of-frequency histogram in the Appendix (Fig.~\ref{fig:binned_frequency}), where GPCRs are grouped into logarithmic bins based on ligand-occurrence counts. This view reveals a pronounced \textit{long-tailed} pattern: most receptors interact with only a handful of ligands, while a relatively small subset of GPCRs accounts for a large fraction of all recorded interactions. This heavy-tailed structure highlights the imbalance that any predictive model must address.

To make the head of the distribution concrete, \autoref{tab:gpcr_detailed_freq_indexed} lists the \textit{Top-10} most frequent receptors (by interaction counts); these highly active receptors dominate the interaction space, whereas the vast majority of GPCRs have far fewer associated ligands (as captured by the histogram).

\subsection{Evaluation protocols and negative sampling}
\label{sec:eval_protocols}
We consider three evaluation scenarios to comprehensively assess GPCR-Filter: (1) a \textbf{random split} for in-distribution performance; (2) an \textbf{intra-target split} to test generalization to new drug combinations over seen targets; and (3) an \textbf{inter-target split} to evaluate cross-target generalization to entirely unseen GPCRs. Task difficulty increases from (1) to (3). All methods operate on the same sequence+SMILES inputs.

\paragraph{Negative sampling (general).}
We evaluated GPCR-Filter under three complementary scenarios including random, intra-target, and inter-target splits to characterize both in-distribution performance and the model’s ability to generalize across ligand and receptor space. All models operated on identical protein-sequence and SMILES inputs.

Because the curated dataset contains only experimentally verified GPCR–ligand pairs, negative examples were generated through systematic negative sampling. For each evaluation setting, we enumerated all possible drug–target combinations, removed known positives, and randomly sampled an equal number of negatives to maintain a 1:1 class balance. Negative sampling was performed independently for each split to avoid information leakage.

In the random split, all positive and sampled negative pairs were pooled, stratified by label, and partitioned into training, validation, and test subsets using an 80/10/10 ratio. This yielded 146,233 training examples, 18,279 validation examples, and 18,280 test examples. The random split measures in-distribution performance when ligands and targets follow the same global distribution across all partitions.

The intra-target split evaluates generalization to unseen ligand combinations for GPCRs that appear during training. Positive pairs were grouped by receptor ID, and negative samples were generated individually for each GPCR by pairing it with all ligands and removing known positives. Targets with fewer than ten positive samples were assigned exclusively to the training set, whereas targets with sufficient data were divided into training, validation, and test subsets using the same 80/10/10 policy. This procedure produced 145,613 training examples, 18,110 validation examples, and 18,225 test examples. This setting preserves target identity across all partitions and challenges models to infer new ligand–target relationships.

The inter-target split tests cross-target generalization to completely unseen GPCRs. All receptor IDs were partitioned into disjoint training and held-out sets at a 9:1 ratio. Positive examples were collected according to this partition, and negative samples were generated separately within the training and held-out target pools. To maintain strict separation of receptor identities, both the validation and test sets were drawn exclusively from the held-out target subset. This resulted in 167,521 training examples, 15,271 validation examples, and 15,271 test examples. The inter-target split provides the most challenging evaluation, requiring models to transfer receptor-level features to entirely novel GPCRs.

\subsection{Baselines and evaluation settings}
GPCR-Filter was compared against two competitive sequence-based DTI architectures under a unified input representation. ConPLex\cite{singh2023contrastive} employs a pretrained protein language model (PLex) alongside contrastive learning to improve generalization to previously unseen proteins and ligands. TransformerCPI2.0\cite{chen2023sequence} integrates a BERT-based protein encoder (TAPE-BERT) with a graph neural network for ligand representations and has shown strong performance in virtual screening and target identification tasks. For all methods, hyperparameters, optimization strategies, and thresholding procedures were matched unless otherwise specified.

\subsection{Binding-pocket visualization procedure}
For interpretability, we analyze test pairs with high-confidence predictions (either top-$N$ per target or $p\ge\tau$, with $\tau=0.5$ defined in the evaluation protocol).
For each selected pair, we extract the \emph{last-layer} ligand$\rightarrow$protein cross-attention from the ligand graph-level token ($t{=}0$) to residue tokens, average over heads (and, if noted, over layers), mask padded positions, L1-normalize the residue-wise scores, and rank residues to obtain a Top-20 list.
We map FASTA indices to PDB residue positions by chain-aware alignment (handling numbering offsets and insertion codes); residues with unknown mapping are ignored.
The binding pocket is defined as residues whose minimum heavy-atom distance to any ligand heavy atom is $<5$\,\AA.
We evaluate agreement using \emph{Pocket hits@20} (the count of pocket residues within the Top-20) and report enrichment relative to a random Top-20 baseline.
For stability analysis, the same procedure is applied to models trained under the random, intra-target, and inter-target splits (quantitative outcomes are presented in Results).

\subsection{\textbf{Virtual screening with GPCR-Filter}}
During the application of GPCR-Filter, molecular docking was first performed using Schrödinger Glide, and the top compounds were selected for GPCR-Filter input. A total of 1,644,833 structures from the ChemDiv database were initially prepared using Schrödinger’s LigPrep. The 5-HT\textsubscript{1A}-SEP-363856 complex (PDB ID: 8W8B), reported in a study of TAAR1/5-HT\textsubscript{1A} dual agonism, \cite{liu2023recognition}, served as the starting point for molecular docking\cite{liu2023recognition}. The 5-HT\textsubscript{1A} receptor was isolated from its G protein complex and processed using the Protein Preparation Wizard in Schrödinger Maestro. Hydrogen atoms were added, disulfide bonds were defined, and residue heteroatom states were assigned using Epik at pH 7.0 ± 2.0. Protonation states were further refined using PROPKA-3\cite{olsson2011propka3}. A receptor grid was generated based on the SEP-363856 binding pocket. High-throughput virtual screening (HTVS) was first applied to identify the top 10\% of compounds, followed by standard precision (SP) docking to retain the top 5\%. Subsequently, 8,705 molecules were screened using GPCR-Filter. Among these, 97 compounds with more than 0.5 predicted binding probabilities were selected for purchase and experimental testing. Due to compound availability, only 52 of these molecules could be obtained and tested experimentally. 

\subsection{\textbf{GloSensor cAMP assay}}
We fused a flag-tag into the N terminal of full-length 5-HT\textsubscript{1A} receptor, and cloned into pcDNA 6.0. 293T cells were cultured in DMEM supplemented with 10\% FBS until reaching 90–95\% confluency and then seeded into 6‐well plates at a density of 450,000 cells per well. Prior to transfection, the medium was exchanged with DMEM containing 10\% charcoal-stripped FBS. For each well, a mixture containing 1.5 μg of the 5-HT$_{1\mathrm{A}}$ receptor plasmid and 1 μg of the cAMP biosensor GloSensor-22F (Promega) construct was combined with 150 μl OPTI-MEM and 7.5 μl of 1 mg/ml polyethylenimine (Yeasen), incubated at room temperature for 15–20 minutes, and then added to the cells.

Twenty hours post-transfection, cells were washed with PBS, detached using 50 μl trypsin per well (1–2 minutes), and the digestion was halted by adding 250 μl DMEM with 10\% charcoal-stripped FBS (Vivacell). After centrifugation, cells were resuspended with HBSS and starved at 37°C for 1 hour. Following starvation, cells were pelleted, resuspended in 500 μl CO₂-independent medium (Gibco), and counted using a 20 μl aliquot. Cells were then mixed with the GloSensor cAMP substrate (Promega) and distributed into 384-well plates, with final cell density 6000/well. After a 1-hour incubation at 37°C, ligands with different fold diluted concentrations was added per well, followed by a 10-minute incubation at room temperature after a brief centrifugation. Finally, Forskolin (final concentration 1 μM) was added to each well, and luminescence was measured by Envision to monitor real-time cAMP responses.

A single-concentration GloSensor-cAMP assay was employed to evaluate GPCR-Filter-prioritized compounds. All 52 available candidates were tested at 30 $\mu$M to identify potential receptor activators. Luminescence (LUM) signals were benchmarked against 5-HT (agonist control) and Forskolin (baseline control), and four ligands (D24, D29, D34, D47) showing significant LUM reductions were selected for subsequent dose–response analysis.

\newpage
\bibliographystyle{unsrt}
\bibliography{Bibliography}

\newpage
\appendix
\clearpage
\onecolumn
\section*{Appendix}
\renewcommand{\thefigure}{S\arabic{figure}}
\setcounter{figure}{0}
\renewcommand{\thetable}{S\arabic{table}}
\setcounter{table}{0}

\begin{center}

\begin{minipage}[t]{0.49\linewidth}
  \centering
  \includegraphics[width=\linewidth]{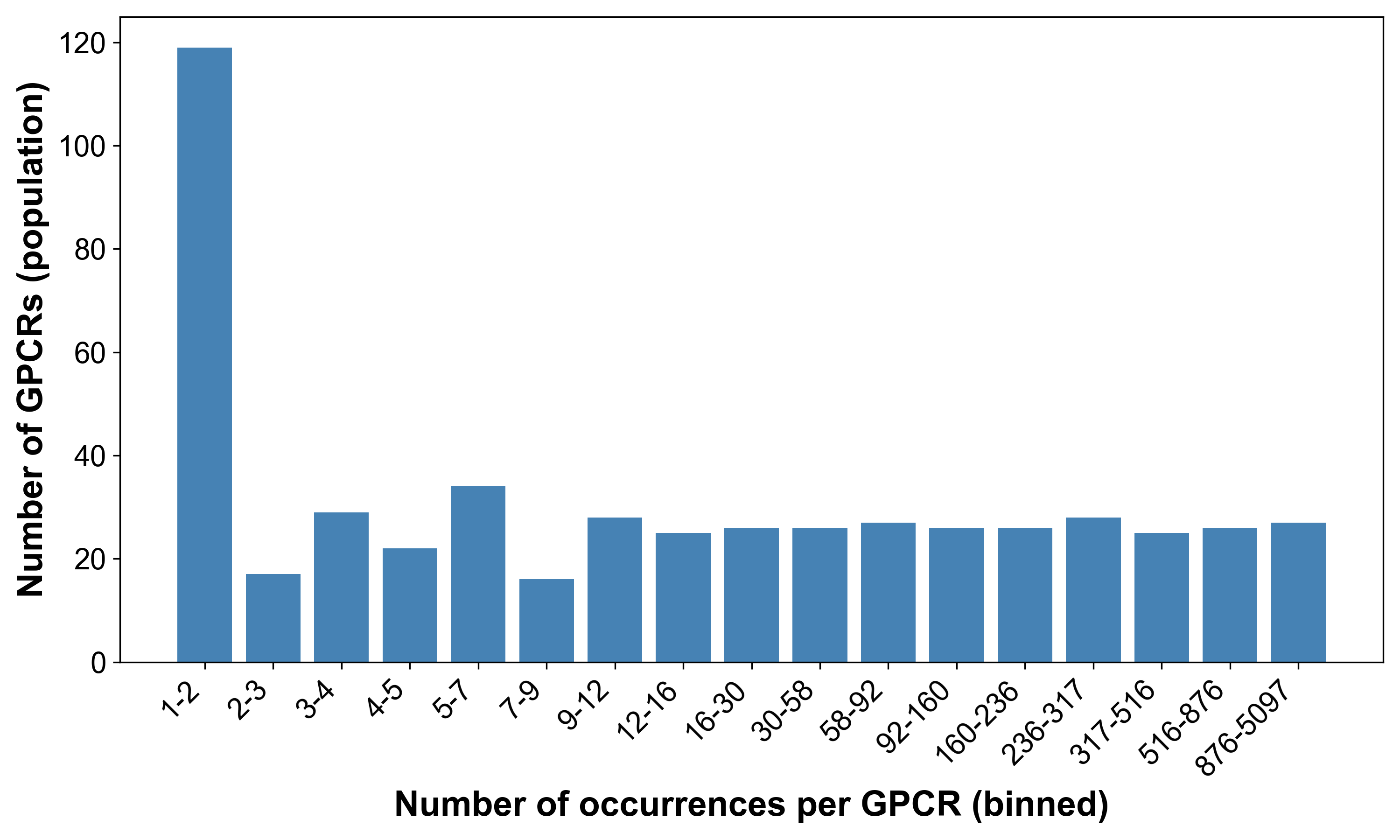}
  \captionof{figure}{\textbf{Distribution of target occurrences per GPCR.}
  Binned frequency-of-frequency histogram summarizing the curated GPCR--ligand dataset.
  The \emph{x}-axis represents the number of ligand occurrences per GPCR (binned), and the \emph{y}-axis represents the number of GPCRs (population) within each bin.
  The pronounced long-tail pattern indicates that while most GPCRs interact with only a few ligands, a small subset of receptors accounts for a disproportionately large number of interactions.}
  \label{fig:binned_frequency}
\end{minipage}
\hfill
\begin{minipage}[t]{0.49\linewidth}
  \centering
  \includegraphics[width=\linewidth]{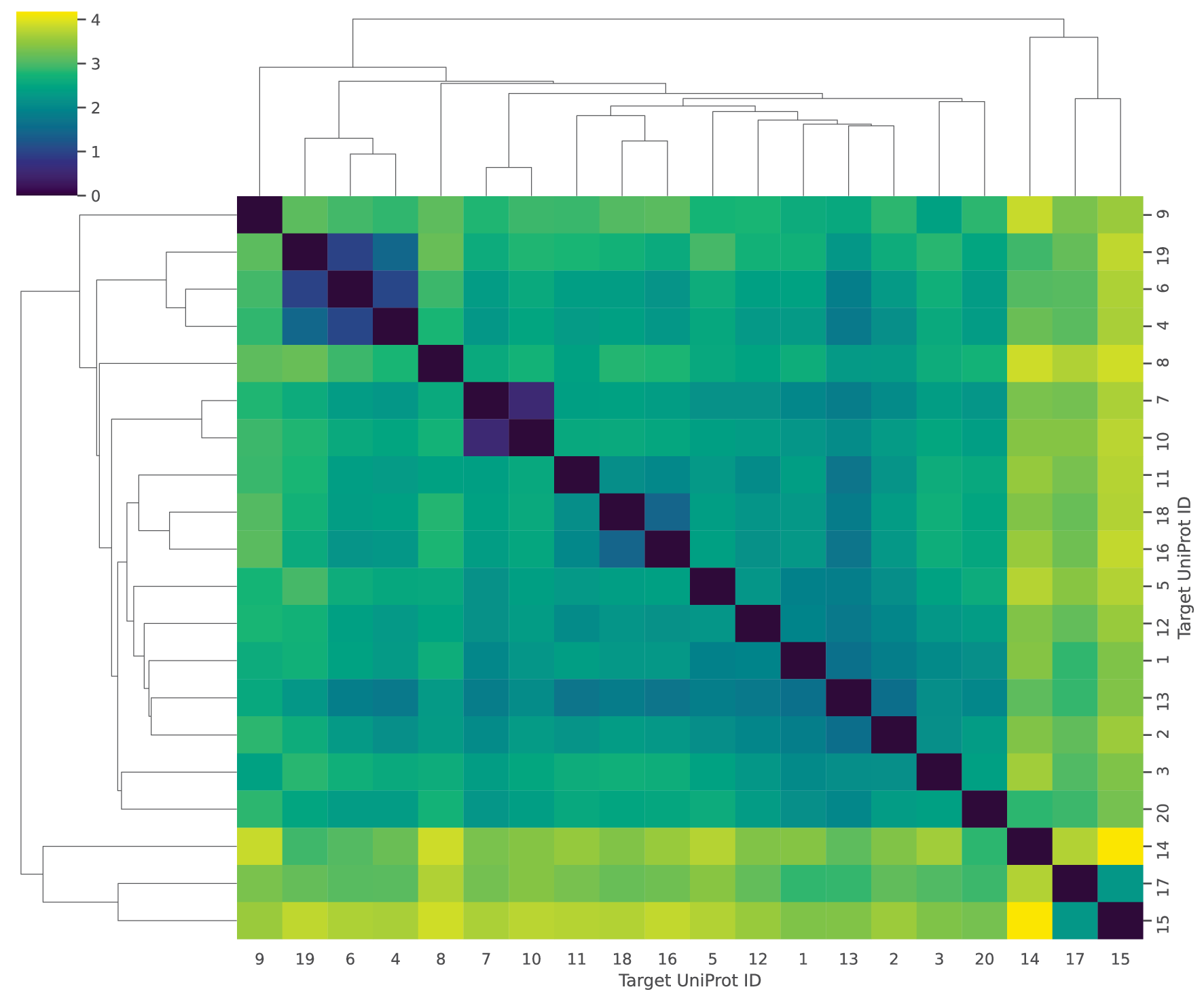}
  \captionof{figure}{\textbf{Hierarchical clustering of GPCRs based on aggregated drug fingerprints.}
  Hierarchical clustering was performed on the GPCR set using drug-fingerprint similarity profiles aggregated over all associated ligands.
  Each cell in the heatmap represents the pairwise similarity between GPCR's ligands, while the dendrograms illustrate hierarchical relationships among targets.
  Closely clustered receptors share comparable ligands, revealing a learnable latent structure within the dataset that supports cross-target generalization in GPCR-Filter.}
  \label{fig:clustering}
\end{minipage}

\vspace{10pt}

\renewcommand{\arraystretch}{1.4}
\captionof{table}{Top-10 most frequent GPCRs in the dataset.}
\label{tab:gpcr_detailed_freq_indexed}

\begin{tabular}{c@{\hspace{30pt}}clcc}
\hline
\textbf{Index} & \textbf{UniProt ID} & \textbf{Protein Name} & \textbf{GPCR Family} & \textbf{Frequency} \\
\hline
1  & Q9HBX9  & Relaxin receptor 1 & GPCR 1 family & 5097 \\
2  & P34972  & Cannabinoid receptor 2 & GPCR 1 family & 3063 \\
3  & P21453  & Sphingosine 1-phosphate receptor 1 & GPCR 1 family & 1982 \\
4  & P41145  & Kappa-type opioid receptor & GPCR 1 family & 1940 \\
5  & P41594  & Metabotropic glutamate receptor 5 & GPCR 3 family & 1917 \\
6  & P35372  & Mu-type opioid receptor & GPCR 1 family & 1799 \\
7  & O43613  & Orexin receptor type 1 & GPCR 1 family & 1735 \\
8  & Q8TDV5  & Glucose-dependent insulinotropic receptor & GPCR 1 family & 1583 \\
9  & O14842  & Free fatty acid receptor 1 & GPCR 1 family & 1479 \\
10 & O43614  & Orexin receptor type 2 & GPCR 1 family & 1444 \\
\hline
\end{tabular}

\end{center}

\newpage
\onecolumn
\end{document}